\documentclass[runningheads]{llncs}

\usepackage[T1]{fontenc}
\usepackage{graphicx}
\usepackage{amsmath}
\usepackage{amssymb}
\usepackage[misc]{ifsym}
\usepackage{comment}

\usepackage{booktabs}
\usepackage{tabularx}
\usepackage{array}
\usepackage{multirow}

\usepackage{subcaption}
\usepackage{enumitem}

\usepackage[ruled,linesnumbered,boxed,shortend,vlined,commentsnumbered]{algorithm2e}

\spnewtheorem{assumption}{Assumption}{\bfseries}{\itshape}

\newcommand{\corr}{(\Letter)}

\setlist[itemize]{leftmargin=*}
\usepackage{hyperref}
\hypersetup{
    colorlinks=true,
    linkcolor=blue,
    filecolor=magenta,
    urlcolor=cyan,
}

\begin{document}
\title{Causal Discovery with Inverted Self-attention for Multivariate Time Series}
\author{Yusen Liu\inst{1} \and
Yong Wang\inst{3} \and
Yifan Yin\inst{1} \and
Tianqing Zhu\inst{4} \and
Xiufeng Liu\inst{2} \corr \and
Huan Huo\inst{1}\corr}
\authorrunning{Liu et al.}

\institute{School of Computer Science, University of Technology Sydney, Sydney, Australia\\
\email{\{Yusen.Liu, Yifan.Yin, Huan.Huo\}@uts.edu.au}
\and
Department of Technology, Management and Economics, Technical University of Denmark, Lyngby, Denmark\\
\email{xiuli@dtu.dk}
\and
TianFu YongXing Laboratory, Chengdu, China\\
\email{ethan-wy@foxmail.com}
\and
Faculty of Data Science, City University of Macau, Macau, China\\
\email{tqzhu@cityu.edu.mo}
}

\maketitle   

\begin{abstract}
Causal discovery in multivariate time series data is challenging due to complex interactions, high dimensionality, and nonlinear dependencies among variables. Existing methods often struggle to capture these complexities, resulting in inaccurate causal structures. To address this issue, we propose a novel framework that leverages self-attention mechanisms within the transformer architecture for causal discovery. Our approach introduces a novel inverted causal self-attention mechanism (CSAM) that emphasizes latent and indirect causal relationships by inverting tokens and inducing sparsity in attention scores, focusing on significant causal interactions and reducing spurious correlations. Additionally, we develop a global causal algorithm to identify global causal links, providing a holistic metric for causal influence, along with a causal verification module to ensure robustness in the identified causal relationships, enhancing the reliability of our framework. Experiments on both linear and nonlinear datasets, along with ablation studies and sensitivity analyses, show that our framework outperforms existing methods, demonstrating its potential for causal discovery in complex multivariate time series.
\keywords{Causal discovery \and Multivariate time series \and Attention mechanisms}
\end{abstract}

\section{Introduction}
Time series analysis is widely applied in fields like economics, environmental science, and healthcare. The temporal dependencies and dynamic nature of time series data create unique challenges, requiring advanced analytical techniques for tasks such as classification and forecasting. Beyond these tasks, understanding causal relationships between variables in multivariate time series holds particular importance, as it not only provides insights into underlying mechanisms but also improves forecasting and classification performance, supporting more informed decision-making across various domains.

However, causal discovery in time series data remains challenging due to the complex interactions between variables over time. A primary difficulty is distinguishing causality from correlation, as temporal dependencies can obscure true causal structures, leading to potential misinterpretations \cite{runge2019inferring}. This challenge is further intensified by the high dimensionality and potential nonlinearity present in multivariate data \cite{runge2018causal}. Traditional methods often struggle to handle these complexities, which can limit research and applications; inaccurate causal links may result in flawed conclusions and less effective interventions, impacting decision-making in critical areas \cite{shih2019temporal}.

Existing methods for causal discovery in time series data, including distance-based \cite{schreiber2000measuring}, index-based \cite{zema2022directed}, and traditional statistical methods \cite{granger1969investigating}, primarily capture linear relationships but are often sensitive to noise and may confuse causation with correlation, limiting their effectiveness in dynamic multivariate contexts \cite{ebert2012causal}. Kernel-based approaches, such as Kernel Granger Causality (KGC) \cite{marinazzo2008kernel}, extend causal discovery to nonlinear relationships. Constraint-based methods, including Time Series Fast Causal Inference (tsFCI) \cite{entner2010causal} and PCMCI \cite{runge2019detecting}, detect causal links even with latent confounders. Score-based methods, such as NOTEARS \cite{zheng2018dags} and DYNOTEARS \cite{pamfil2020dynotears}, offer additional tools for uncovering causal structures in multivariate time series, addressing some limitations of traditional methods. Despite these advances, there has been limited exploration of deep learning’s potential for causal discovery \cite{nauta2019causal}, especially given its capacity to model complex, high-dimensional data.

To further address high dimensionality and nonlinearity challenges in multivariate time series data, we propose a novel framework that leverages an inverted self-attention mechanism for causal discovery. First, we propose an inverted causal self-attention mechanism (CSAM) that highlights latent and indirect causal relationships. Specifically, we apply Sparsemax in the attention score matrix, inducing sparsity to focus on significant causal interactions and reduce spurious correlations. Then, we develop a global algorithm to integrate attention scores across model components, providing a metric for causal influence that captures complex causal structures in time series data. This approach enables the model to handle nonlinear interactions while remaining robust to noise and confounders, improving interpretability and reducing false positives. Finally, we integrate Permutation Importance (PI) \cite{pereira2022covered} to verify that identified causal relationships are robust, enhancing the reliability of our causal discovery pipeline. In summary, the main contributions of this paper are: 
\begin{itemize}
    \item We propose a novel framework that leverages an inverted causal self-attention mechanism for causal discovery, enabling the capture of complex, nonlinear causal relationships in multivariate time series.
    \item We introduce a sparsity-inducing modification to the attention score matrix, improving interpretability and reducing spurious correlations.
    \item We develop a global causal algorithm to identify global causal links, along with a causal verification module to ensure robustness in the identified relationships.
    \item We validate our framework through experiments on high-dimensional and complex time series data, with ablation studies and sensitivity analyses demonstrating the contribution of each component.
\end{itemize}

The structure of this paper is as follows: Section~\ref{sec:relatedwork} reviews related work on causal discovery in time series and attention mechanisms. Section~\ref{sec:definitions} introduces the definitions and assumptions of this paper. Section~\ref{sec:methods} details our proposed methodology. Section~\ref{sec:exp} presents the experiments and results. Finally, Section~\ref{sec:con} concludes the paper and suggests future research directions.

\section{Related Work}
\label{sec:relatedwork}
\paragraph{Causal Discovery for Time Series.} 
Causal discovery in time series data aims to identify causal relationships, with Granger causality as a foundational method. It posits that if past values of a time series \(Y\) improve the prediction of another series \(X\), then \(Y\) Granger-causes \(X\) \cite{granger1969investigating}. Building on this, KGC \cite{marinazzo2008kernel} extends Granger's framework to nonlinear relationships via kernel methods. Distance-based methods \cite{schreiber2000measuring} also capture nonlinear relationships but are sensitive to noise and require careful parameter tuning. Additionally, constraint-based approaches like tsFCI \cite{entner2010causal} detect causal links even with latent confounders. PCMCI \cite{runge2019detecting} combines the PC algorithm \cite{spirtes1991algorithm} with momentary conditional independence tests for efficient causal structure identification. Score-based methods, such as NOTEARS \cite{zheng2018dags} and DYNOTEARS \cite{pamfil2020dynotears}, use continuous optimization to capture temporal and nonlinear dependencies. Further, \cite{tank2021neural,nauta2019causal} leverage neural networks to infer causality directly. Together, these methods offer a range of tools for analyzing causal relationships in time series.
\paragraph{Attention Mechanisms.} Attention mechanisms enable models to focus on relevant data segments and are widely used in fields like NLP and computer vision. First introduced by \cite{bahdanau2014neural} for machine translation, attention improved both model performance and interpretability. The Transformer architecture by \cite{vaswani2017attention}, relying entirely on attention, set new benchmarks in NLP. In time series analysis, attention mechanisms have shown strong potential in models like Informer \cite{zhou2021informer} and Crossformer \cite{zhang2022crossformer}. Attention-based models have also been applied in causal discovery, with \cite{nauta2019causal} incorporating attention into convolutional neural networks to improve causal effect estimation. In contrast to these approaches, we propose a novel CSAM to identify latent and nonlinear causal relationships in multivariate time series data, facilitating a deeper understanding of dynamic interactions.

\section{Definitions and Assumptions}
\label{sec:definitions}
\begin{definition}{\textbf{\textup{(Multivariate Time Series)}}} 
A multivariate time series is defined as an $M$-dimensional series $\mathbf{X} = (\mathbf{X}^1, \mathbf{X}^2, \ldots, \mathbf{X}^M)$ of length $T$, where each component $\mathbf{X}^i$ represents an individual time series.
\end{definition}
\vspace{-8pt}
\begin{definition}{\textbf{\textup{(Causal Graph)}}}\label{def:causal_graph}
Given a multivariate time series $\mathbf{X}$, its causal graph is a directed acyclic graph (DAG) $G = (V,E)$, where $V = \{1, 2, \dots, M\}$ represents the set of variables (time series), and $E$ is the set of directed edges. An edge $(i, j) \in E$ exists if and only if $\mathbf{X}^i$ Granger-causes $\mathbf{X}^j$.
\end{definition}
\vspace{-8pt}
\begin{assumption}{\textbf{(Causal Markov)}}\label{asmp:causal_markov}
The causal Markov condition states that each variable is conditionally independent of non-effects given its direct causes, focusing the analysis on direct causal dependencies.
\end{assumption}
\vspace{-8pt}
\begin{assumption}{\textbf{(Faithfulness)}}\label{asmp:faithfulness}
Faithfulness assumes that any conditional independencies in the data reflect the true causal structure, ensuring that observed independencies correspond to actual causal relationships.
\end{assumption}
\vspace{-8pt}
\begin{assumption}{\textbf{(Strict Causal Ordering)}}\label{asmp:causal_ordering}
Strict temporal ordering in the causal graph implies that if $\mathbf{X}^i$ causes $\mathbf{X}^j$, then $i < j$, disallowing contemporaneous causation and simplifying temporal dependency analysis.
\end{assumption}
\vspace{-8pt}
\begin{assumption}{\textbf{(Bounded Dependencies)}}\label{asmp:bounded_dependencies}
For each edge $(i,j) \in E$, a constant $\epsilon > 0$ exists such that the minimum absolute difference between the prediction error of $\mathbf{X}^j$ with and without $\mathbf{X}^i$ as a predictor exceeds $\epsilon$ for sufficiently large $T$, ensuring a minimum "signal strength" for true causal dependencies.
\end{assumption}

\section{Methods}
\label{sec:methods}
Following our proposed definitions and assumptions, we introduce a novel framework that systematically addresses the identification of Granger causality networks in multivariate time series $\mathbf{X}$. The model aims to delineate causal interactions among series components through matrix $A$. The full time series $\mathbf{X}_t$ is input to a causal self-attention module (CSAM), which differs from traditional self-attention in token handling and internal dynamics. This module’s output feeds into a transformer, generating predictions for each series $\mathbf{X}^i_t$ and identifying causal links with other series during training. Granger causality is then derived from $\mathbf{X}_t$ through a global algorithm and verification strategy, as detailed in Sections \ref{sec:global} and \ref{sec:verify}.


\subsection{Causal Self-attention Mechanism}
\label{sec:csam}
\begin{figure}[t!]
    \centering
    \includegraphics[width=\textwidth]{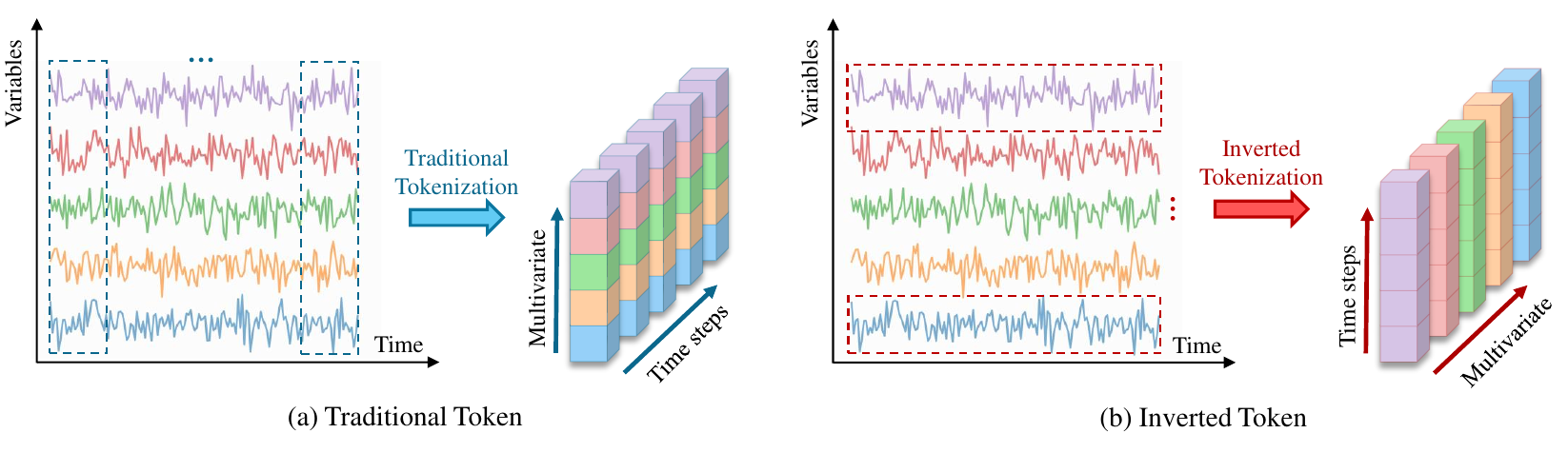}
    \caption{Comparison of token representations in (a) traditional self-attention and (b) the proposed causal self-attention mechanism.\label{fig2}}
\end{figure}

Transformers effectively capture complex dependencies in sequential data through self-attention mechanisms. To leverage this for causal discovery, we introduce an inverted self-attention mechanism, CSAM, designed to identify causal relationships in multivariate time series data. Traditional self-attention computes attention scores based on the affinity between positions in the input sequence, formulated as:
\begin{equation}
    \text{Attention}(Q, K, V) = \text{softmax}\left(\frac{QK^T}{\sqrt{d_k}}\right)V
\end{equation}
where $Q$, $K$, and $V$ represent the matrices for queries, keys, and values, respectively, and $d_k$ is the dimensionality of the keys. In multivariate time series analysis, each token $X = \{x_1, x_2, x_3, \ldots\}$ represents a combination of different variables, as shown in Fig.~\ref{fig2}(a). Each token corresponds to a single time point, and a sequence of tokens $\{X_1, X_2, X_3, \ldots\}$ represents consecutive time points. Using these tokens in the self-attention mechanism allows for the computation of inter-temporal relationships.

Contrastingly, in Granger causality analysis, the focus is on identifying relationships among different time series rather than across time intervals. To achieve this, we invert the token representation. As shown in Fig.~\ref{fig2}(b), each inverted token $\hat{X}$ consists of sequential observations within a single time series, $\hat{X} = \{x^1, x^2, x^3, \ldots\}$, representing one complete time series. A collection of these inverted tokens, $\{\hat{X_1}, \hat{X_2}, \hat{X_3}, \ldots\}$, represents multiple time series. Using these inverted tokens in the self-attention framework allows for the analysis of cross-series relationships, aligning with Granger causality objectives. This approach maintains the temporal sequence within each series while enabling a detailed exploration of causal dynamics across the multivariate time series.

\begin{figure}[t!]
    \centering
    \includegraphics[width=0.9\textwidth]{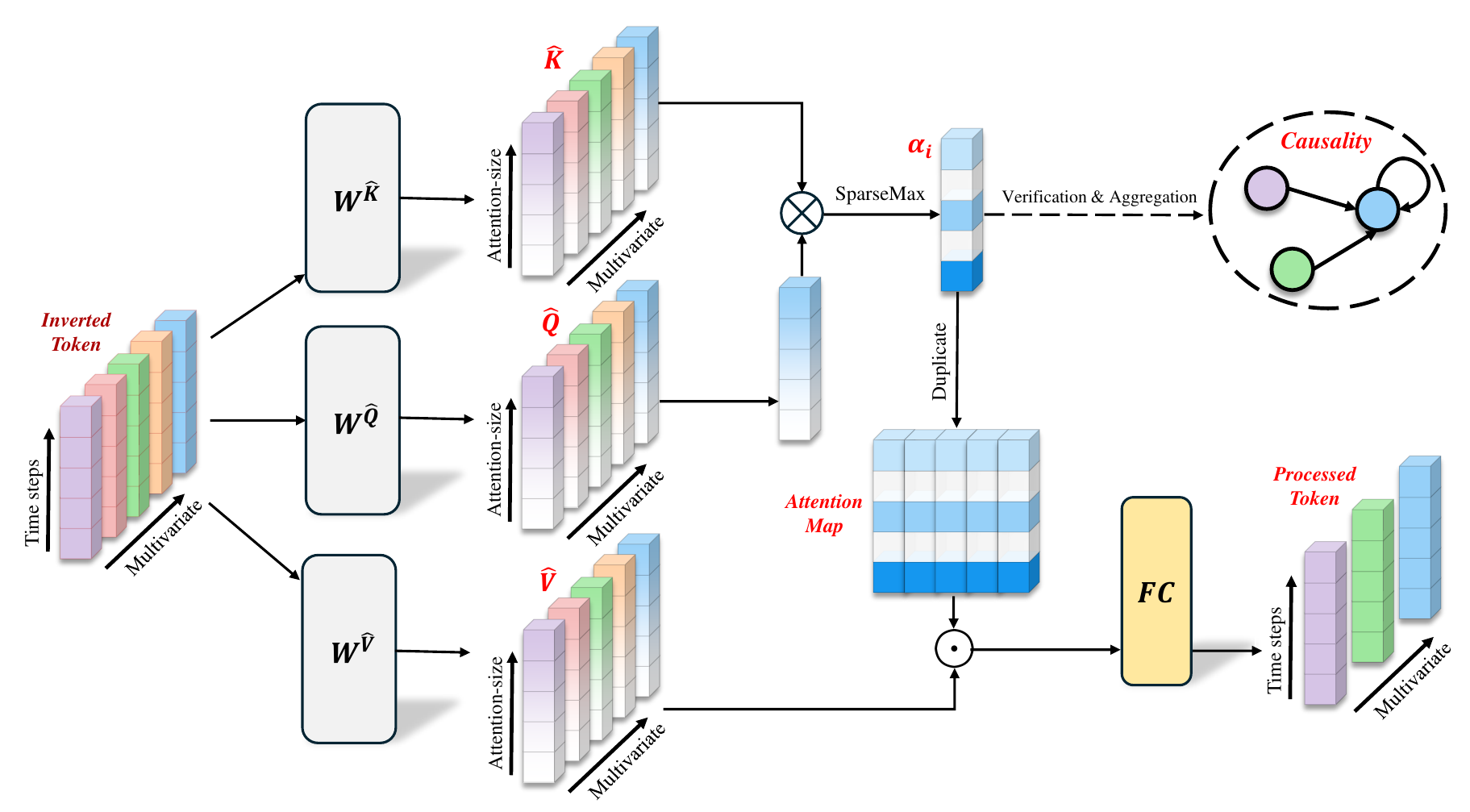}
    \caption{Illustration of the process of our causal self-attention mechanism (CSAM).}
\label{fig3}
\end{figure}

After transposing the tokens, we apply the CSAM to reveal Granger causality among the time series, as illustrated in Fig.~\ref{fig3}. The inverted tokens are initially transformed through multiplication with the weight matrices $W^{\hat{k}}$, $W^{\hat{q}}$, and $W^{\hat{v}}$, producing the key ($\hat{K}$), query ($\hat{Q}$), and value ($\hat{V}$) matrices:
\begin{equation}
    \hat{K} = \hat{X} \cdot W^{\hat{k}} , \quad \hat{Q} = \hat{X} \cdot W^{\hat{q}} , \quad \hat{V} = \hat{X} \cdot W^{\hat{v}}  
\end{equation}
It is imperative to note that, diverging from the conventional dimensions observed in standard transformers, which are typically indexed by \( {K} \in \mathbb{R}^{N \times d_k} \), \( {Q} \in \mathbb{R}^{N \times d_q} \), \( {V} \in \mathbb{R}^{N \times d_v} \), the dimensions of our key ($\hat{K}$), query ($\hat{Q}$), and value ($\hat{V}$) matrices are structured as \( \hat{K} \in \mathbb{R}^{T \times d_k} \), \( \hat{Q} \in \mathbb{R}^{T \times d_q} \), \( \hat{V} \in \mathbb{R}^{T \times d_v} \), Where $N$ stands for the number of variables and $T$ stands for the time steps. Focusing on a specific time series designated by $i$, we extract the $i$-th row from the $\hat{Q}$ matrix and engage in a matrix multiplication with the transposed $\hat{K^T}$ matrix. This operation is instrumental in deriving the initial attention scores $\vec{e_i}$ that encapsulate potential causal influences exerted by all time series on the series of interest $i$:
\begin{equation}
    \vec{e_i} = \frac{\hat{Q}_i \times \hat{K}^T} {\sqrt{d_k}}
\end{equation}

To enhance the interpretability of the resultant weight vector, we apply the \textit{SparseMax} function \cite{martins2016softmax}. This transformation imparts sparsity to the vector, thereby facilitating a more straightforward interpretation by accentuating the most salient causal relationships and diminishing the noise from negligible interactions. The \textit{SparseMax} function is a differentiable alternative to the softmax function, designed to provide sparser probabilities. Formally, the \textit{SparseMax} function for a vector $\mathbf{z} \in \mathbb{R}^d$ is defined as the solution to the following optimization problem:
\begin{equation}
    \textit{SparseMax}(\mathbf{z}) = \underset{\mathbf{p} \in \Delta^{d-1}}{\textit{argmin}} \left\| \mathbf{p} - \mathbf{z} \right\|^2
\end{equation}
where $\Delta^{d-1}$ denotes the $(d-1)$-dimensional simplex, i.e., the set of $\mathbf{p} \in \mathbb{R}^d$ such that $\sum_{i=1}^d p_i = 1$ and $p_i \geq 0$ for all $i$. The \textit{SparseMax} operation projects the input vector $\mathbf{z}$ onto the simplex, resulting in a sparse probability distribution where many elements can be exactly zero.
\begin{equation}
    \vec{\alpha}_i = \textit{SparseMax }(\vec{e_i})
\end{equation}
After applying \textit{SparseMax}, the non-zero elements in the sparse vectors $\vec{\alpha_i}$ are added to the causal tensor $\mathbf{C}_{ijt}$ as potential causes of the $i$-variables, where $t$ denotes different tokens. Additionally, $\vec{\alpha_i}$ is duplicated across the $d_v$ dimension to form an attention map $\mathbf{A_i}$ for the CSAM. The attention map represents the refined causal influence structure, with each element indicating the degree of influence one time series has on another within the multivariate context.
\begin{equation}
    \mathbf{A_i} = \textit{Replicate}(\vec{\alpha}_i, d_v) = \begin{bmatrix} \vec{\alpha}_i & \vec{\alpha}_i & \cdots & \vec{\alpha}_i \end{bmatrix}
\end{equation}

The final stage performs an element-wise multiplication (Hadamard product) between the attention map $\mathbf{A_i}$ and the $\mathbf{V}$ matrix, encapsulating the weighted significance of each value element based on the derived causal relationships. This product is then fed into a fully connected layer to produce the final processed token $\hat{X}'$, which serves as input to the transformer model.
\begin{equation}
    \hat{X}' = \text{FC}(\mathbf{A_i} \odot \mathbf{V})
\end{equation}

\subsection{Global Causal Algorithm}
\label{sec:global}
Algorithm \ref{alg:global} presents the global causal algorithm, which systematically identifies potential causal links in the dataset based on causal strengths derived from the attention mechanism. This approach aggregates causal information from each token during training to output a set of potential causes $\mathbf{P}$. Specifically, the attention mechanism is applied to each token during training, generating a causal tensor $\mathbf{C}_{ijt}$, where $t$ represents different tokens. The algorithm then calculates the causal strengths $H_{ij}$ between each pair of time series $i$ and $j$ by summing $\mathbf{C}_{ijt}$ across all tokens:
\begin{equation}
H_{ij} = \sum_t \mathbf{C}_{ij}^t
\end{equation}
Next, the algorithm sorts the causal strengths ${H_{ij}}$ in descending order and selects a threshold $\tau$ based on a predefined parameter $k$, where $\tau$ is set as the $k$-th largest value in the sorted list.
\begin{equation}
\tau = \text{sorted}({H_{ij}})[k]
\end{equation}
We select the threshold $\tau$ following the approach in \cite{nauta2019causal}. If the causal strength $H_{ij}$ between any two time series $i$ and $j$ meets or exceeds $\tau$, the algorithm infers a causal relationship from $i$ to $j$ and includes it in the set of potential causes $\mathbf{P}$:
\begin{equation}
    \mathbf{P} = \{(i \rightarrow j) \mid H_{ij} \geq \tau\}
\end{equation}
\begin{algorithm}[t!]
\small 
\SetAlgoNlRelativeSize{-1} 
\SetAlgoNlRelativeSize{-1} 
\SetAlgoInsideSkip{2pt} 
\caption{Global Causal Algorithm\label{alg:global}}
\KwData{Causality tensor $\mathbf{C}$, order parameter $k$}
\KwResult{potential causes set $\mathbf{P}$}
\SetKwFunction{FMain}{GlobalCausality}
\SetKwProg{Fn}{Function}{:}{}
\Fn{\FMain{$C$, $k$}}{
    \For{$i = 1$ \KwTo $N$}{
        \For{$j = 1$ \KwTo $N$}{
            $H_{ij} \leftarrow \sum_t \mathbf{C}_{ij}^t$\;
        }
    }
    $hs \leftarrow \textbf{sorted}(\{H_{ij}\})$\;
    $\tau \leftarrow hs[k]$\;
    $\mathbf{P} \leftarrow \emptyset$\;
    \For{$i = 1$ \KwTo $N$}{
        \For{$j = 1$ \KwTo $N$}{
            \If{$H_{ij} \geq \tau$}{
                $\mathbf{P} \leftarrow \mathbf{P} \cup \{(i \rightarrow j)\}$\;
            }
        }
    }
    \KwRet $\mathbf{P}$\;
}
\end{algorithm}

\subsection{Causal Verification}
\label{sec:verify}
Our methodology uses Permutation Importance (PI) \cite{pereira2022covered} to validate causal relationships.
PI quantifies the effect of variable reordering on prediction error by shuffling the values of a candidate cause variable, disrupting potential causal pathways and assessing the significance of the temporal structure.

Given a neural network $\mathcal{N}_j$ trained on the complete dataset to predict the target time series $\mathbf{X}_j$, we denote the baseline loss on the original data as $\mathcal{L}_G$. For each potential causal factor $\mathbf{X}_i \in \mathbf{P}$, we create a permuted dataset by randomly rearranging the values of $\mathbf{X}_i$ while keeping the rest of the data unchanged. This permutation preserves the original data distribution, allowing us to re-evaluate the prediction for $\mathbf{X}_j$ without retraining the network. The perturbed dataset produces a new loss metric $\mathcal{L}_I$. A significant increase in loss from $\mathcal{L}_G$ to $\mathcal{L}_I$ suggests a strong causal influence of $\mathbf{X}_i$ on $\mathbf{X}_j$. If the increase is minor, it indicates that the temporal ordering of $\mathbf{X}_i$ is not crucial for predicting $\mathbf{X}_j$, questioning its causal role. The set of validated causes $\mathbf{C}_j$ for each time series is determined through this process, forming the basis for constructing the Granger causality diagram $G(V, E)$, where $V$ represents the time series nodes and $E$ the directed edges validated by the causal sets.

\section{Experiments}
\label{sec:exp}
\subsection{Experimental Setup}
Our implementation was based on Python 3.8.18 and PyTorch 1.12.1, with experiments conducted on a server equipped with two NVIDIA A40 GPUs. We employed Mean Squared Error as the loss function and utilized the Adam optimizer with a learning rate of 0.001, modified by a scheduler with a step size of 10 and gamma of 0.98. Hyper parameters were set through preliminary tuning, with a batch size of 32 for 500 epochs. To reduce overfitting, dropout was applied. Input and output window sizes were configured to 10 and 1, respectively, while both causal attention and transformer modules employed 8 heads and 3 layers.
\paragraph{\textbf{\textup{Datasets}}}
Our experiments used linear and non-linear multivariate time series datasets, as summarized below.
\textit{Linear Datasets:}
1) \textbf{Hénon Maps:} Chaotic time series data from 6 interconnected Hénon maps, totaling 2,048 data points for training \cite{kugiumtzis2013direct}.
2) \textbf{FINANCE:} A 25-variable financial time series dataset \cite{kleinberg2013causality}, utilized for evaluating causal discovery algorithms.
\textit{Non-linear Datasets:}
1) \textbf{Lorenz-96 Model:} A 10-variable atmospheric dynamics model generating 2,048 samples \cite{lorenz1996predictability}.
2) \textbf{fMRI:} Simulated BOLD signals derived from dynamic causal modeling for fMRI data, with 1,200 samples for analysis \cite{yan2020reconstructing}.

\paragraph{\textbf{\textup{Baseline Methods}}}
We compared our method with six standard models in Granger causal discovery: \textbf{BGranger} \cite{granger1969investigating}, \textbf{KGC} \cite{marinazzo2008kernel}, \textbf{tsFCI} \cite{entner2010causal}, \textbf{TCDF} \cite{nauta2019causal}, \textbf{PCMCI} \cite{runge2019detecting}, and \textbf{DYNOTEARS} \cite{pamfil2020dynotears}. These methods represent diverse approaches to causal discovery. BGranger provides a foundational benchmark by testing if past values of one variable improve predictions of another. Constraint-based methods like tsFCI and PCMCI use conditional independence tests to infer causal structures. KGC, a kernel-based method, extends Granger causality to non-linear relationships. TCDF enhances inference with attention mechanisms, and DYNOTEARS, an optimization-based method, applies continuous optimization for dynamic causal discovery.
\paragraph{\textbf{\textup{Evaluation Metrics}}}
To evaluate performance, we use \textbf{Precision} $(P)$, \textbf{Recall} $(R)$, and \textbf{F1-Score} $(F_1)$ to measure accuracy. Precision represents the fraction of correctly identified causal relationships among those inferred, Recall indicates the fraction of actual causal relationships detected, and F1-Score provides a balanced measure between Precision and Recall.

\begin{table*}[t!]
\vspace{-5pt}
\centering
\caption{Performance comparison of the proposed method and baselines.\label{tab:combined}}
\scriptsize %
\begin{tabularx}{\textwidth}{@{}l|*{3}{>{\centering\arraybackslash}X}|*{3}{>{\centering\arraybackslash}X}|*{3}{>{\centering\arraybackslash}X}|*{3}{>{\centering\arraybackslash}X}@{}}
\toprule
& \multicolumn{3}{c|}{Henon} & \multicolumn{3}{c|}{Finance} & \multicolumn{3}{c|}{Lorenz-96} & \multicolumn{3}{c}{fMRI} \\
\midrule
Method & Prec. & Rec. & F1 & Prec. & Rec. & F1 & Prec. & Rec. & F1 & Prec. & Rec. & F1 \\
\midrule
BGGranger & 0.238 & 0.455 & 0.312 & 0.041 & 0.108 & 0.059 & 0.295 & 0.307 & 0.283 & 0.250 & 0.140 & 0.182 \\
KGC        & 0.412 & 0.636 & 0.500 & 0.223 & 0.208 & 0.215 & 0.466 & 0.406 & 0.548 & 0.437 & 0.333 & 0.378 \\
tsFCI      & 0.643 & \textbf{0.818} & 0.720 & 0.424 & 0.301 & 0.356 & 0.583 & 0.389 & 0.467 & 0.474 & 0.429 & 0.450 \\
TCDF       & 0.875 & 0.636 & 0.737 & 0.521 & 0.307 & 0.387 & 0.735 & 0.694 & 0.714 & 0.800 & 0.762 & 0.780 \\
PCMCI      & 0.889 & 0.727 & 0.799 & 0.624 & 0.501 & 0.556 & 0.703 & \textbf{0.722} & 0.712 & 0.704 &\textbf{0.904} & 0.792 \\
DYNOTEARS      & 0.845 & 0.677 & 0.752 & 0.653 & 0.364 & 0.467 & 0.699 & 0.697 & 0.698 & 0.547 & 0.442 & 0.489 \\
\textbf{Ours}       & \textbf{0.900} & \textbf{0.818} & \textbf{0.857} & \textbf{0.673} & \textbf{0.505} & \textbf{0.586} & \textbf{0.862} & 0.684 & \textbf{0.769} & \textbf{0.818} & 0.857 & \textbf{0.837} \\
\bottomrule
\end{tabularx}
\end{table*}

\begin{figure}[t!]
    \centering
    \includegraphics[width=1\textwidth]{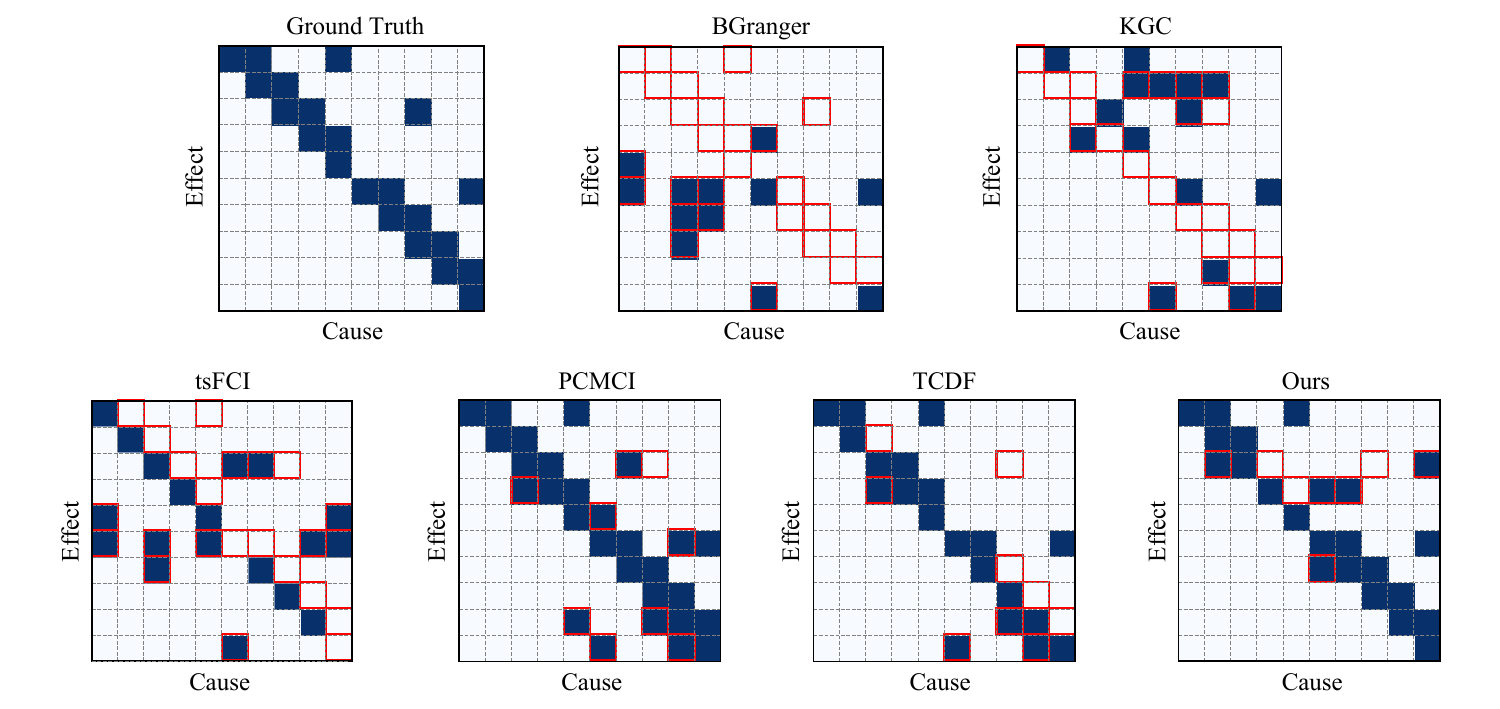}
    \vspace{-10pt}
    \caption{Comparison of causality matrices from various methods on the fMRI dataset, where dark blue indicates true causality and red highlights errors.
    \label{fig:visual_comparison}}
\end{figure}

\subsection{Overall Performance}
The quantitative results in Table~\ref{tab:combined} demonstrate the effectiveness of our causal discovery framework across different datasets. Our method consistently achieves high precision scores, from 0.673 for the FINANCE dataset to 0.900 for the Hénon dataset, indicating its strong ability to minimize false-positive causal inferences. Furthermore, our approach maintains competitive recall scores, showing its capacity to capture a significant portion of the ground-truth causal relationships. Notably, our method outperforms existing state-of-the-art techniques on the non-linear Lorenz-96 and fMRI datasets, achieving the highest F1-scores of 0.769 and 0.837, respectively. This performance underscores the robustness of our framework in uncovering complex causal structures within non-linear systems, a key requirement for real-world applications.

The visual comparison in Fig.~\ref{fig:visual_comparison} further supports the quantitative results. The inferred causality matrix from our method closely matches the ground truth, with minimal false positives and negatives. This accuracy is due to the combined effects of the proposed CSAM, the sparsity-inducing modifications, and the global integration algorithm. CSAM captures complex, non-linear causal dependencies by leveraging self-attention mechanisms within the transformer architecture. The sparsity-inducing modifications enhance the reliability of identified causal relationships by reducing spurious correlations and noise. Finally, the global integration algorithm provides a holistic metric for causal influence, facilitating the discovery of both direct and indirect causal links while accounting for temporal dynamics and inter-dependencies among variables.

\begin{figure}[t!]
    \centering
    \begin{subfigure}{.32\textwidth}
        \centering
        \includegraphics[width=\linewidth]{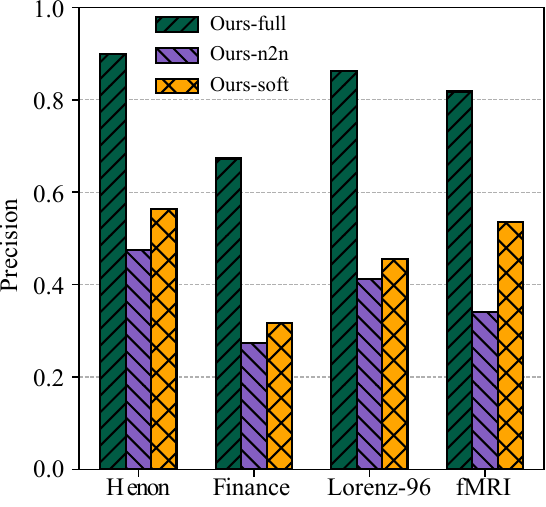}
        \caption{Precision}
        \label{fig:ablation_precision}
    \end{subfigure}%
    \hfill
    \begin{subfigure}{.32\textwidth}
        \centering
        \includegraphics[width=\linewidth]{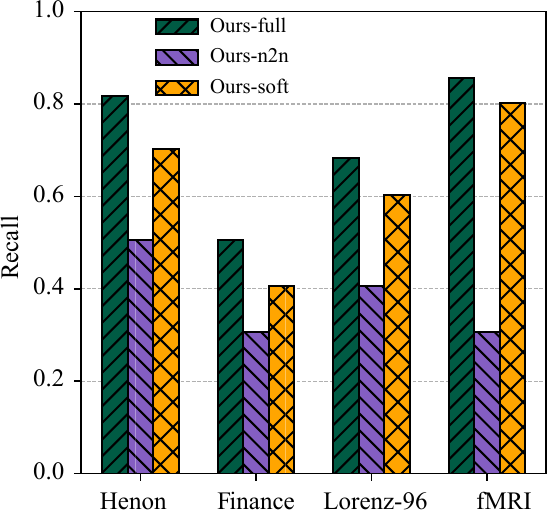}
        \caption{Recall}
        \label{fig:ablation_recall}
    \end{subfigure}%
    \hfill
    \begin{subfigure}{.32\textwidth}
        \centering
        \includegraphics[width=\linewidth]{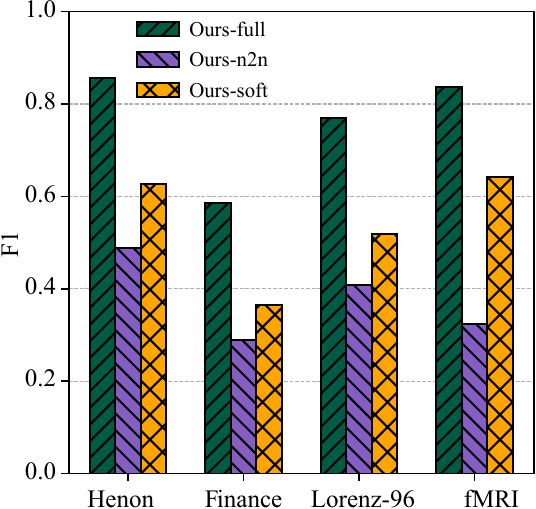}
        \caption{F1-Score}
        \label{fig:ablation_f1}
    \end{subfigure}
    \caption{Ablation study results across datasets.\label{fig:ablation_study}}
\end{figure}

\subsection{Ablation Study}
To evaluate the contributions of our model's components, we conducted an ablation study focusing on the CSAM and sparsity-inducing mechanism. Our study compares the performance of the complete model (\texttt{Ours-full}) with two variants: \texttt{Ours-n2n} and \texttt{Ours-soft}. The \texttt{Ours-n2n} variant, which predicts all elements of the multivariate time series simultaneously, shows reduced precision compared to \texttt{Ours-full}, indicating that parallel prediction disperses focus and weakens causal detection, as shown in Fig.~\ref{fig:ablation_precision}. The \texttt{Ours-soft} variant, replacing SparseMax with Softmax for attention scoring, also shows a precision drop across datasets, as Softmax fails to create the desired sparsity, reducing specificity in causal inference. As seen in Fig.~\ref{fig:ablation_recall}, these variants affect recall as well. While \texttt{Ours-n2n} performs similarly to \texttt{Ours-full} in some datasets, it underperforms in complex cases like Lorenz-96 and fMRI. \texttt{Ours-soft} also exhibits inconsistencies, especially in datasets where sparsity is key to accurate causal detection. The F1-scores in Fig.~\ref{fig:ablation_f1} confirm that \texttt{Ours-full} outperforms the variants, demonstrating the importance of both CSAM and sparsity-inducing mechanisms.

\subsection{Sensitivity Analysis}
\begin{minipage}{0.55\textwidth}
    The CSAM, rooted in the transformer framework, inherently relies on multiple attention heads and blocks, both of which are crucial hyper-parameters in the model's architecture. In our parameter analysis, the number of Attention Heads was varied among 2, 4, 8, and 16, with 8 set as the default value. Similarly, the number of Attention Blocks was adjusted to 1, 2, 3, and 4, with 3 being the default.
    \end{minipage}%
    \hfill
    \begin{minipage}{0.4\textwidth}
        \centering
        \captionof{table}{Sensitivity analysis.}
        \vspace{-5pt}
        \label{tab:ParameterStudy}
        \resizebox{\textwidth}{!}{
        \footnotesize 
        \begin{tabular}{*{5}{c}}
            \toprule
           Parameters & No. &  Prec. & Rec. & F1\\
            \midrule
            \multirow{4}{*}{{Attention Head}}  & 2 & 0.783 & 0.698 & 0.738 \\
            & 4 & 0.769 & 0.802 & 0.785 \\
            & 8 & \textbf{0.818} & \textbf{0.857} & \textbf{0.837} \\
            & 16 & 0.747 & 0.737 & 0.742 \\
            \midrule
            \multirow{4}{*}{{Attention Block}}  & 1 & 0.779 & 0.796 & 0.788 \\
            & 2 & 0.764 & 0.834 & 0.814 \\
            & 3 & \textbf{0.818} & \textbf{0.857} & \textbf{0.837} \\
            & 4 & 0.799 & 0.854 & 0.826 \\
            \bottomrule
        \end{tabular}
        }
    \end{minipage}

\textit{Number of Attention Heads:} In the CSAM, the number of attention heads controls how many attention mechanisms are applied to the input data. Each head processes the sequence separately, allowing the model to capture various dependencies. As shown in Table~\ref{tab:ParameterStudy}, increasing attention heads to 16 helps capture more complex dependencies but may cause over-fitting. Conversely, reducing heads to 2 simplifies the model, potentially losing important information. 

\textit{Number of Attention Blocks:} Attention blocks determine the depth of attention applied across the time series data, with each block adding a layer of processing to refine the model's understanding. Table~\ref{tab:ParameterStudy} shows that increasing blocks to 4 enhances the model’s ability to capture complex dependencies but risks over-fitting, while reducing blocks to 1 may fail to capture sufficient details.

\section{Conclusion and Future Work}
\label{sec:con}
This study introduces a novel framework for causal discovery in multivariate time series data. Our proposed inverted causal self-attention mechanism (CSAM) provides a unique token representation to analyze causal relationships. By integrating a global causal algorithm with causal verification, the framework effectively handles spurious correlations. Experiments on four benchmarks demonstrate that our method outperforms existing approaches. For future research, we aim to extend the method to complex datasets across different domains and explore approaches focused on identifying true causal frequencies and causal lags.

\begin{credits}
\subsubsection{\ackname} This work was supported by the China Scholarships Council (Grant No. 202208410132), and the Tianfu Yongxing Laboratory Organized Research Project Funding (No. 2023CXXM14).
\end{credits}

\bibliographystyle{splncs04}
\bibliography{reference}
\end{document}